\documentclass[pmlr]{jmlr}

\RequirePackage{graphicx}
 \usepackage{booktabs}
\usepackage{longtable}
\makeatletter
\def\set@curr@file#1{\def\@curr@file{#1}} 
\makeatother
\usepackage[load-configurations=version-1]{siunitx} 

\jmlrvolume{252}
\jmlryear{2024}
\jmlrworkshop{Machine Learning for Healthcare}

\usepackage{textcomp}
\usepackage{multirow}
\usepackage{caption}

\title[Fairness of Normative Models]{To which reference class do you belong? Measuring racial fairness of reference classes with normative modeling}

\author{\Name{Saige Rutherford}
       \Email{saige.rutherford@donders.ru.nl}\\ 
       \addr Department of Cognitive Neuroscience\\
       Radboud University Medical Center\\
       Nijmegen, the Netherlands
       \AND
       \Name{Thomas Wolfers}
       \Email{dr.thomas.wolfers@gmail.com}\\ 
       \addr Department of Psychiatry\\
       University of Tuebingen\\
       Tuebingen, Germany
       \AND
       \Name{Charlotte Fraza}
       \Email{charlotte.fraza@donders.ru.nl}\\ 
       \addr Department of Cognitive Neuroscience\\
       Radboud University Medical Center\\
       Nijmegen, the Netherlands
       \AND
       \Name{Nathaniel G. Harnett}
       \Email{nharnett@mclean.harvard.edu}\\ 
       \addr Department of Psychiatry\\
       Harvard Medical School\\
       Boston, MA, USA
       \AND
       \Name{Christian F. Beckmann}
       \Email{christian.beckmann@donders.ru.nl}\\ 
       \addr Department of Cognitive Neuroscience\\
       Radboud University Medical Center\\
       Nijmegen, the Netherlands
       \AND
       \Name{Henricus G. Ruhe}
       \Email{eric.ruhe@radboudumc.nl}\\ 
       \addr Department of Psychiatry\\
       Radboud University Medical Center\\
       Nijmegen, the Netherlands
       \AND
       \Name{Andre F. Marquand}
       \Email{andre.marquand@donders.ru.nl}\\
       \addr Department of Cognitive Neuroscience\\
       Radboud University Medical Center \\
       Nijmegen, the Netherlands
       }

\begin{document}

\maketitle

\begin{abstract}
    Reference classes in healthcare establish healthy norms, such as pediatric growth charts of height and weight, and are used to chart deviations from these norms which represent potential clinical risk. How the demographics of the reference class influence clinical interpretation of deviations is unknown. Using normative modeling, a method for building reference classes, we evaluate the fairness (racial bias) in reference models of structural brain images that are widely used in psychiatry and neurology. We test whether including “race” in the model creates fairer models. We predict self-reported race using the deviation scores from three different reference class normative models, to better understand bias in an integrated, multivariate sense. Across all of these tasks, we uncover racial disparities that are not easily addressed with existing data or commonly used modeling techniques. Our work suggests that deviations from the norm could be due to demographic mismatch with the reference class, and assigning clinical meaning to these deviations should be done with caution. Our approach also suggests that acquiring more representative samples is an urgent research priority. 
\end{abstract}

\section{Introduction}

Reference classes can be used to define health and disease in medicine and also to estimate patient risk. Determining the probability of a patient possessing a particular attribute, such as a disease or prognosis, is critical for risk screening and treatment planning. Through direct inference (that is, making inferences about the individual from the population \cite{thorn_two_2012}, probabilities for individual patients can be adjusted based on estimates derived from reference classes to which the patient belongs. Determining which reference class is most useful can be challenging when a patient fits into multiple reference classes, each with differing risk probabilities. This is known as the reference class problem  \cite{hajek_reference_2007,wallmann_four_2017,venn_logic_1888,reichenbach_theory_1949}. Due to the difficult nature of the reference class problem, most demographic information (excluding perhaps age and sex) is ignored when calculating clinical risk. While ignoring the reference class problem may allow easier group identification, doing so can also lead to worse (i.e., less accurate, more disparate\footnote{Throughout this work, we use the terms fairness, bias, and disparity interchangeably. We consider them all to refer to testing for equal model performance across different self-reported racial categories.}) risk prediction in high-stake settings \cite{suriyakumar_when_2023,khor_racial_2023,pfohl_empirical_2021,zink_race_2023}. For example, consider two hypothetical patients with renal cell carcinoma with brain metastases, John, who is a 34-year-old, white male, unmarried with no family close by, living in rural Michigan, low-income, exercises regularly, non-drinker, who occasionally smokes, and Jim, who is a 68-year-old black male, married, living in New York City, high-income, does not exercise regularly, drinks heavily, and does not smoke. Each of these demographics creates its own reference class that may contradict each other (e.g., regular exercise and smoking). When calculating the chance of survival for each man, the Graded Prognostic Assessment tool is used \cite{sperduto_survival_2020}. Even if both men have the same score, their respective reference classes (socioeconomic status, family support, exercise levels, alcohol intake, smoker/non-smoker) will likely influence their survival probabilities.

Normative modeling is a framework for building reference class models and is, therefore, ideal for studying the reference class problem. Normative modeling has been applied across many healthcare contexts, including the most well-known use case in pediatrics - growth charting of height, weight, and head circumference \cite{borghi_construction_2006}. Fields of medicine that involve heterogeneous disease categories and complex measurements of biology, such as psychiatry \cite{wolfers_mapping_2018,zabihi_dissecting_2019,lv_individual_2020,elad_improving_2021} and neurology \cite{bhome_neuroimaging_2023,verdi_beyond_2021,verdi_personalising_2023,italinna_detecting_2022}, have begun to use normative modeling to move away from standard case vs. control methods which mainly consider group mean effects \cite{marquand_beyond_2016,marquand_conceptualizing_2019,rutherford_normative_2022,bethlehem_brain_2022} towards individualized, or precision, medicine. In addition to clinical applications, there has been considerable methodological improvements in normative modeling, including cross-sectional vs. longitudinal modeling \cite{buckova_using_2024, dibiase_mapping_2023}, handling site effects \cite{bayer_accommodating_2021,kia_closing_2022,kia_federated_2021}, predictive modeling using outputs of normative models \cite{rutherford_evidence_2023}, heterogeneity quantification \cite{nunes_definition_2020,nunes_measuring_2020,nunes_we_2020}, extreme value statistics \cite{fraza_extremes_2022}, and modeling non-Gaussianity \cite{boer_non-gaussian_2022}.

\begin{figure}[t]
  \centering 
  \includegraphics[width=\textwidth]{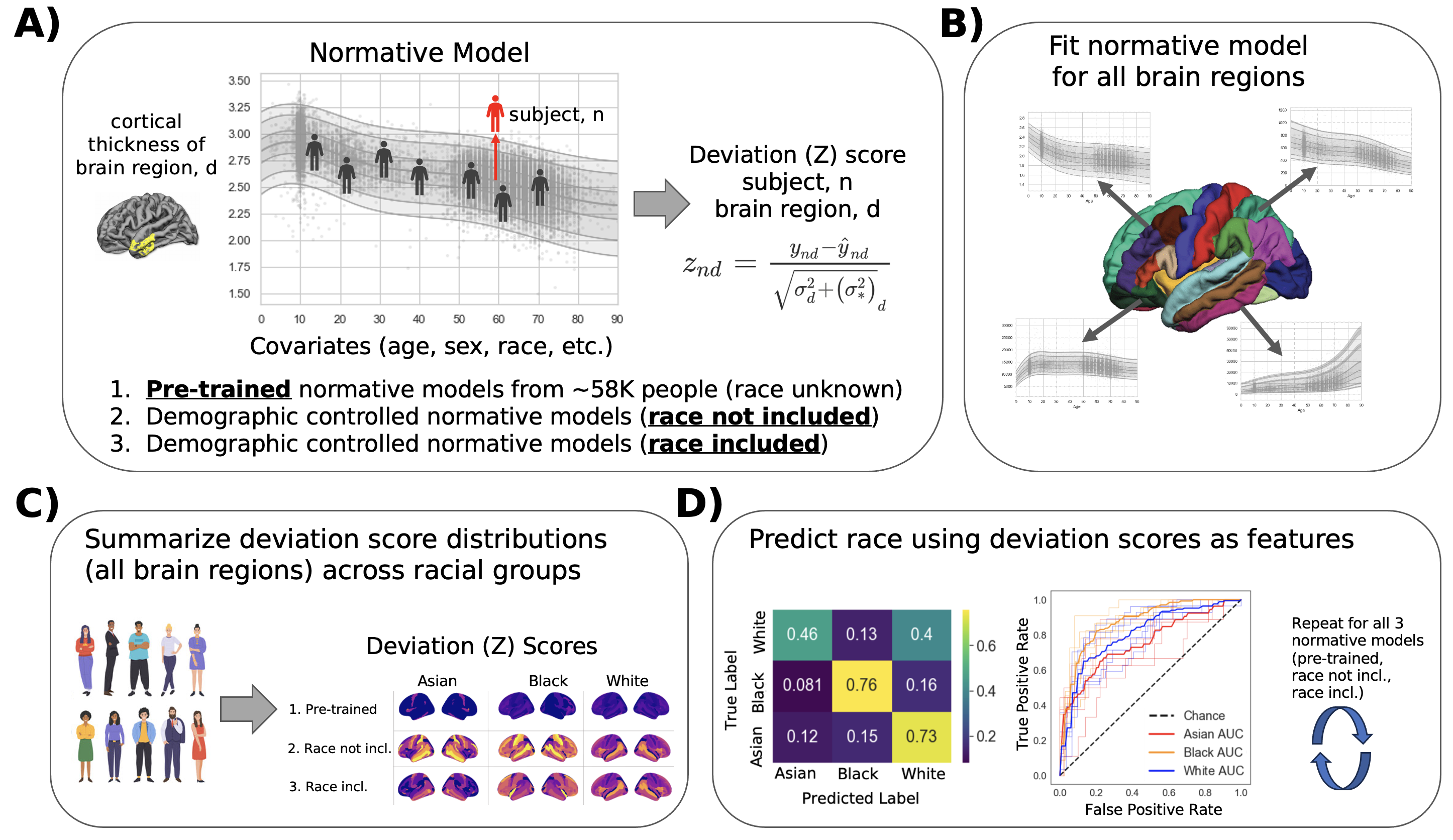} 
  \caption{Overview of analysis workflow. A) Normative models of brain structure were used to generate deviation scores. Three normative models were fit (pre-trained, race not included, and race included) representing two different reference classes and two sets of covariates. B) Normative models were estimated for all regions in the Destrieux atlas \cite{destrieux_automatic_2010}, a commonly used anatomical brain parcellation. C) The effect of self-reported race on the distribution of normative modeling deviation scores was quantified across all three normative models. D) Self-reported race was predicted using normative modeling deviation scores as features.}
  \label{fig:figure1} 
\end{figure}

An important concept in normative modeling is untangling various sources of heterogeneity (variability) within the population. Normative models aim to distinguish between ‘healthy’ and clinical variation, as well as other factors like demographic or confounding variables (e.g., site/scanner). Typically, 'healthy' variability is assumed to fall within the 95th percentiles, while deviations associated with illness are assumed to fall within the outer percentiles (Figure \ref{fig:figure1}A). However, it is also possible that deviations in the outer percentiles could relate to demographics rather than being clinically relevant, which is usually not empirically tested. Despite the increasing use of normative modeling in both machine learning and healthcare domains, the fairness implications of these models have not been thoroughly investigated. Considerations for determining who should be included in the reference group often revolve around data availability and increasing sample sizes. However, an essential unresolved question is how the demographics of this reference class affect the interpretation of deviation scores. Before addressing concerns regarding fairness in reference classes as operationalized in normative models, it is important to first quantify them. 

\subsection*{Generalizable Insights about Machine Learning in the Context of Healthcare}

\begin{itemize}
    \item We first quantify the racial bias in existing (pre-trained) normative models of structural brain images that are widely used in psychiatry and neurology \cite{rutherford_charting_2022}. Racial demographics of the reference class used to train these existing normative models are unknown because many of the samples used for the training set did not collect or share race or ethnicity data. We probe these models, using a sample where race is known, by summarizing deviations and residual errors across Asian, Black, and White individuals and testing for group differences (Figure \ref{fig:figure1}A-C).  
    \item Next, we train new normative models using reference class data where race information is available during training. In this setting, we test whether including self-reported race variables into the normative model in an identical cohort (race included and race not included) creates fairer models, by summarizing deviations and residual errors across Asian, Black, and White individuals and testing for group differences (Figure \ref{fig:figure1}A-C).
    \item Finally, we use the deviation scores from three different reference class normative models (pre-trained, race included, race not included) to predict self-reported race. Normative models are fit in a univariate manner, meaning one brain region is predicted per model, and there are models fit across many brain regions. The race prediction uses deviation scores from all brain regions combined into the same model to better understand bias in an integrated, multivariate sense (Figure \ref{fig:figure1}D). 
\end{itemize}

\section{Background and Problem Formulation}

\subsection{Fairness in Machine Learning for Healthcare Setting}

Prior work on fairness in machine learning considers three criteria: independence, separation, and sufficiency \cite{barocas_fairness_2023}. For independence (often called demographic parity), the predictions should be statistically independent of the sensitive attribute (e.g. race) \cite{corbett-davies_algorithmic_2017,kamiran_classifying_2009}. In separation (often called equalized odds), sensitive attributes are statistically independent of the prediction given the ground truth \cite{hardt_equality_2016}. For sufficiency (often called predictive parity), sensitive attributes are statistically independent of the ground truth given the prediction (similar to group calibration) \cite{berk_fairness_2017,chouldechova_fair_2016}. There are also three stages of the machine learning lifecycle where fairness concerns are often addressed: pre-processing, training time, and post-processing. Data acquisition is also an obvious source of bias. However, this stage is often disregarded as it is considered outside control of the machine learning community who are typically performing secondary data analysis and not directly involved in data collection \cite{paullada_data_2021,sambasivan_everyone_2021,roh_survey_2021}. In pre-processing, the training data are modified such that the information correlated to the sensitive attribute is removed. For training time fairness, there is typically some penalty added to the loss function (e.g., joint optimization of a fairness metric and the loss function). Post-processing fairness is often performed by applying a transformation to the model output to reduce unfairness. There are numerous fairness metrics used in machine learning for healthcare and they vary depending on the context. There are too many fairness metrics to review in this work but we refer to recent work on this topic in the classification \cite{franklin_ontology_2022} and regression setting \cite{franklin_ontology_2023}.

Across all of these definitions of fairness and implementation stages, it becomes clear that much of the research on fairness in machine learning for healthcare is trying to solve fairness by eliminating bias with some new method or metric. An underlying assumption seems to be that models can be made perfectly fair with the right optimization or algorithm. Similarly, research on the reference class problem is also often focused on developing more “correct” reference classes through various approaches such as the digital twin \cite{turab_comprehensive_2023,bruynseels_digital_2018} or improved modeling of overlapping probabilities \cite{wallmann_four_2017}. Less attention is given to simply measuring rather than eliminating bias in widely used datasets, reference classes, and predictive models. Quantifying bias in predictive healthcare models rather than focusing on eliminating it might be as straightforward as creating guidelines for their use based on performance across different groups. 

\subsubsection{Racial Fairness}

When considering the fairness of machine learning models with respect to racial identity, extensive research has been conducted on challenging the inclusion of race as a predictor (race corrections\footnote{We use "race correction" as it is standard terminology \cite{ioannidis_recalibrating_2021, vyas_hidden_2020}, however, we disagree with this name because it is impossible to correct properly if you have insufficient data to define an alternative reference class.} \cite{vyas_hidden_2020,cerdena_race-based_2020,borrell_race_2021,ioannidis_recalibrating_2021,khor_racial_2023,yang_evaluating_2023,suriyakumar_when_2023}. Racial corrections in medicine can be traced back to the practice of using white male bodies as the reference, or norm, against which other bodies and physiological functions are measured. Due to biased data collection, the white male norm principally reflects data availability, as has been the case in many biomedical domains (e.g. genetics or brain imaging), but it does not mean that this reference class should be the gold standard of health against which all others are measured. The consensus on race correction suggests avoiding a binary approach (always perform or always avoid), but rather in every context carefully considering the rationale behind its use to filter out scientifically debunked beliefs about biological differences between racial groups \cite{roberts_abolish_2021}, and calls for further research into the multifaceted interactions among genetic ancestry, race, racism, socioeconomic status, and environmental factors \cite{vyas_hidden_2020}. \textbf{It is important to explicitly state that race is not simply a phenotype related to skin color but is necessarily concomitant with a complex and interactive set of influences from society and the environment.}

There has been important work documenting existing racial bias in clinical predictive models, for example in explainable AI methods \cite{balagopalan_road_2022}, clinical record de-identification \cite{xiao_name_2024}, and underdiagnosis in chest X-ray pathology \cite{seyyed-kalantari_underdiagnosis_2021}. Studies have shown that race can be predicted from clinical notes \cite{adam_write_2022} and medical images (chest x-rays, chest CTs, mammograms, and spine x-rays) \cite{gichoya_ai_2022}. In work that directly predicted race from medical images, detection of race was not due to proxies for race (body-mass index, disease distribution, breast density) and race could be predicted from all anatomical regions and frequency spectrums of the images \cite{gichoya_ai_2022}. Another topic of work has looked at why racial disparities exist in clinical predictive models. This includes studying label misspecification \cite{pierson_algorithmic_2021, obermeyer_dissecting_2019}, undertesting rates of certain minority groups \cite{chang_disparate_2022, wu_collecting_2023,akpinar_impact_2024}, and testing whether disparities are due to differences in outcome frequency, feature distributions or feature-outcome relationships \cite{movva_coarse_2023}. The work on underdiagnosis in chest X-ray pathology \cite{seyyed-kalantari_underdiagnosis_2021} also opened further conversations on understanding the sources of racial bias, including population, prevalence, and annotation shifts \cite{bernhardt_potential_2022, mukherjee_confounding_2022, seyyed-kalantari_reply_2022}.

\subsection{Normative Model Setting}

 The normative model setting is a natural approach for studying fairness of the reference class problem because it can be seen as a statistical model of a particular reference class. When selecting a specific normative model, we essentially define a concrete reference group, which can then be used to quantitatively evaluate fairness. It is nevertheless somewhat distinctive from many machine learning in healthcare settings. Therefore, we briefly describe normative modeling at a conceptual level and in section \ref{sec:normativemodelestimation}, we formulate the normative model mathematically. A normative model is most often used to model population variation in biological measurements across a particular reference class. For example, pediatric growth charts are a well-known example that plot centiles of variation in height or weight as a function of age.  Normative modeling can also be thought of as a feature generating step (i.e., creating features for downstream clinical decisions) whereas most machine learning in healthcare models operate directly at the clinical decision-making step (e.g., predicting clinical outcomes such as hospitalization, critical outcomes (ICU transfer), or re-admission \cite{movva_coarse_2023, xie_benchmarking_2022}. However, it is important to recognize that if the features are biased, then it is likely that downstream algorithms will be too \cite{chang_disparate_2022}.  
More concretely, we can employ a pre-trained predictive regression model (Figure \ref{fig:figure1}A) estimated on the basis of a set of binary and continuous features, $X_{nd}$ from $n$ subjects, to predict a continuous response variable, $y_d$, representing a biological measurement (the thickness of the cerebral cortex in brain region $d$). In the setting of neuroimaging, many normative models are fit for each biological feature (e.g. for each anatomical brain region derived from brain imaging). The model is typically a population reference model that pools data from many different study sites. Covariates $X_{nd}$ are variables that are considered to be important in predicting the biological response variable, yet in practice are constrained to include only the overlapping available variables across sites (typically age and sex for most neuroimaging studies). A central goal of these ‘population’ reference class normative models is to be openly shared and to be able to transfer the models to new sites or datasets (e.g. which might contain individuals with a given medical condition). At some of these transfer sites, there are additional variables relating to demographics (race) that we would like to use to test the model’s fairness across different racial groups (Figure \ref{fig:figure1}C). These demographic variables are only available during model transfer and are typically inaccessible before or during training, because, surprisingly, many of the samples that were pooled for training did not collect or share race or ethnicity data. The outputs of fitting a normative model are statistics that measure the overall fit of the model and a set of deviation scores ($Z_{nd}$-scores), which is as an expression of how far a given person (e.g. a patient) deviates from the reference class population mean (Figure \ref{fig:figure1}A) and is often interpreted as risk of an adverse clinical event \cite{alexander-bloch_abnormal_2014,marquand_conceptualizing_2019}. Clinical labels are often used to make statements such as ‘patients in group A have a greater number of extreme deviations than controls.’ \cite{lv_individual_2020,pinaya_using_2021}.

\section{Cohort}

\subsection{Cohort Selection and Inclusion Criteria}

We used data from two publicly available datasets of neuroimaging and phenotypic data for our experiments, namely the Human Connectome Project (HCP) \cite{glasser_minimal_2013} and the UK Biobank (UKB) \cite{alfaro-almagro_image_2018}. Inclusion criteria for all samples used in our analysis participants having basic demographic information (age, sex, and race) and a high quality (defined in section \ref{sec:featureextraction}) T1-weighted MRI volume. Importantly, both HCP and UKB collected self-reported race variables. Due to the low number of people in some racial groups, we combined UKB categories into Asian (Indian, Pakistani, Asian or Asian British, Bangladeshi, Chinese), Black (Caribbean, African, Black or Black British), and White (White, British, Irish). In HCP, categories included: White, Black or African American, Native American/Native Alaskan, and Asian/Native Hawaiian/Other Pacific Islander. The HCP Native American/Native Alaskan category was not large enough to stand alone as a category, did not fit into other categories, and was not included in our analyses. We did not include UKB or HCP subjects with unknown, unreported, or mixed race. We acknowledge that the Asian group is heterogeneous and quite different across HCP and UKB samples. UKB is more representative of South Asian population (i.e., India, Bangladeshi), whereas HCP is likely more representative of east Asian populations (i.e., Chinese). We also acknowledge that these coarse representations of race are not ideal and we expand on this further in the discussion. Sample demographics of each data set are described in Tables \ref{tab:table1} and \ref{tab:table2}.

\subsection{Train/test split - Normative models}

Slightly different train/test splits were used for the pre-trained models (Table \ref{tab:table1}) and the normative models trained in this work (race-included and race-not-included, Table \ref{tab:table2}). For the pre-trained models, we followed the train/test split of the original paper (split-half, stratified by site) \cite{rutherford_charting_2022}. For the race-included and race-not-included models, after learning of the racial bias in the pre-trained models, we had a hypothesis that the pre-trained models’ training set was primarily composed of white individuals. However, because we did not have access to race information for individuals in the training set of the pre-trained models, we could not assess this. This is when we decided to train our own models in samples where we had access to self-reported race for all individuals in the train/test set. We wanted to test if having a primarily white training set led to the same racial bias we observed in the pre-trained models. So, we stratified the train/test split (Table \ref{tab:table2}) on race and only included a very small number of Black and Asian individuals in the training set. Due to the very low sample size of Black and Asian individuals in HCP and UKB, we wanted to have most of these people in the test set to increase our power. However, to learn the race term during training, we had to include a small subset of these groups in the training sets (2\% Asian, 5\% Black in HCP and 0.2\% Asian and 0.2\% Black in UKB). 

\begin{table}
\centering
\caption{Test set demographics of pre-trained normative models. Demographics of the training set are unknown because many of the samples that were pooled for the training set did not collect or share race or ethnicity data. A = Asian, B = Black, W = White.}
\label{tab:table1}
\begin{tabular}{c|c|c}
~ & HCP test & UKB test \\ 
\hline
N & 533 & 13416 \\ 
\hline
Sex (F\%, M\%) & 53.7\%, 46.3\% & 52.2\%, 47.8\% \\ 
\hline
Age (M, S.D) & 28.9, 3.6 & 63.5, 7.5 \\ 
\hline
Race (A\%, B\%, W\%) & 5.5\%, 15\%, 79.5\% & 1\%, 1\%, 98\% \\
\hline
\end{tabular}
\end{table}

\begin{table}
\centering
\caption{Train and test set demographics of race not included, and race included normative models. Models were fit separately for HCP and UKB data sets. The train/test split was the same across the race not included and the race-included models. A = Asian, B = Black, W = White.}
\label{tab:table2}
\setlength{\tabcolsep}{4pt}
\begin{tabular}{l|l|l|l|l}
~ & HCP train & HCP test & UKB train & UKB test \\ 
\hline
N & 710 & 353 & 29553 & 1472 \\ 
\hline
Sex (F\%, M\%) & 54\%, 46\% & 56\%, 44\% & 53\%, 47\% & 50\%, 50\% \\ 
\hline
Age (M, S.D) & 29.0, 3.5 & 28.6, 4.0 & 63.7, 7.5 & 62.2, 7.8 \\ 
\hline
Race (A\%, B\%, W\%) & 2\%, 5\%, 93\% & 15\%, 38\%, 47\% & 0.2\%, 0.2\%, 99.6\% & 19\%, 19\%, 62\% \\
\hline
\end{tabular}
\end{table}

\subsection{Feature Extraction}
\label{sec:featureextraction}

Freesurfer image analysis software (version 6.0) was used to reconstruct the volumetric MRI neuroimages into a surface (Figure \ref{fig:figure1}B), which better represents the brain’s gyri and sulci folding patterns. The boundary between white matter and cortical gray matter is mapped during Freesurfer’s surface reconstruction which allows cortical thickness (measured in millimeters) to be calculated. Cortical thickness values were extracted for all brain regions in the Destrieux parcellation \cite{destrieux_automatic_2010}, a commonly used brain atlas spanning 148 brain regions (74 left hemisphere, 74 right hemisphere). Further technical details of these procedures are described in prior publications \cite{fischl_measuring_2000,fischl_whole_2002}. We adopted an automated quality control procedure that quantifies image quality based on the Freesurfer Euler Characteristic, which has been shown to be an excellent proxy for manual labeling of scan quality \cite{monereo-sanchez_quality_2021,rosen_quantitative_2018} and is the most important feature in automated scan quality classifiers \cite{klapwijk_qoala-t_2019}. 

\section{Methods}

\subsection{Normative Model Estimation}
\label{sec:normativemodelestimation}

Normative models were estimated using Bayesian Linear Regression with likelihood warping to predict cortical thickness from a vector of covariates \cite{fraza_warped_2021}. For each $d$ brain region of interest, $y_d$ is predicted as:

\begin{equation}
\label{eq:normmodel}
y_d = w^T \phi(x)+\epsilon
\end{equation}

Where $w^T$ is the estimated weight vector, $\phi(x)$ is a basis expansion of the of covariate vector $x$, consisting of a B-spline basis expansion (cubic spline with 5 evenly spaced knots) to model non-linear effects of age, and $\epsilon = \eta(0,\beta)$ a Gaussian noise distribution with mean zero and noise precision term $\beta$ (the inverse variance). A likelihood warping approach \cite{rios_compositionally-warped_2019,snelson_warped_2003} was used to model non-Gaussian effects. This involves applying a bijective nonlinear warping function to the non-Gaussian response variables to map them to a Gaussian latent space where inference can be performed in closed form. We used a ‘sinarcsinsh’ warping function, equivalent to the SHASH distribution used in the generalized additive modeling literature \cite{jones_sinh-arcsinh_2009}. A fast numerical optimization algorithm was used to optimize hyperparameters (L-BFGS) and computational complexity of hyperparameter optimization was controlled by minimizing the negative log likelihood.  Deviation scores (Z-scores) are calculated in the latent Gaussian space for the $n_{th}$ subject, and $d_{th}$ brain area, in the test set as:

\begin{equation}
\label{eq:zscore}
Z_{nd} = \frac{y_{nd} - \hat{y}_{nd}}{\sqrt{\sigma^2_{d} + (\sigma_*^2)_d}}
\end{equation}

Where $y_{nd}$ is the true response, $\hat{y}_nd$ is the predicted mean,  $\sigma^2_{d}$ is the estimated noise variance (reflecting uncertainty in the data), and $(\sigma_*^2 )_d$ is the variance attributed to modeling uncertainty. 

Error is calculated as the model residual:

\begin{equation}
\label{eq:error}
E_{nd} = y_{nd} - \hat{y}_{nd}
\end{equation}

Model fit was evaluated by explained variance, mean squared log-loss, skew, and kurtosis (Appendix Figure \ref{fig:supfig1}, Appendix Figure \ref{fig:supfig2}).

Three normative models were used in this work, all of them following the same modeling and evaluation procedure described above but differing in the training and testing data. First, we wanted to quantify racial bias in existing normative models that are publicly \hyperlink{https://pcnportal.dccn.nl/}{available} and are currently being used in practice in the fields of psychiatry and neurology. These pre-trained models were trained on large (~58K people) data sets and limited demographic information was used during training (age and sex). The racial demographics of the reference class used in the pre-trained models are unknown. Next, to study the effects of race in normative models more carefully, we trained two normative models using a sample where the racial demographics are known. We trained one model with race included as ‘one hot’ dummy variable encoded predictors and one model without race included. The training and testing samples were held constant across both models, the only difference is that one model included race as a predictor, and one did not. To summarize, we wanted to study the effects of race in three different settings, namely: (i) where we first transfer from a large pre-trained model derived from a sample where we cannot determine the demographic distribution (ii) train from scratch in a more carefully controlled subsample where racial information is present and (iii) train from scratch but additionally accounting for the effect of race during training using the same data as and training/test split used in (ii). The differences between these three normative models are summarized in Table \ref{tab:table3}. All analysis code is available on GitHub.\footnote{\url{https://github.com/saigerutherford/nm_demographics/tree/main}}

\begin{table}[!h]
\captionsetup{justification=raggedright, singlelinecheck=false}
\centering
\caption{Summary of Normative Models.}
\label{tab:table3}
\begin{tabular}{l|l|l|l}
Model Name & Model Equation & Training Data & Testing Data \\ 
\hline
Pre-trained & \begin{tabular}[c]{@{}l@{}}Y (brain region) =\\~Age + Sex + Site\\~\end{tabular} & \begin{tabular}[c]{@{}l@{}}\textasciitilde{}58,000 people, \\race is unknown.\end{tabular} & \begin{tabular}[c]{@{}l@{}}HCP \& UKB (table \ref{tab:table1}), \\race is known and used \\post-hoc to quantify bias.\end{tabular} \\ 
\hline
Race not included & \begin{tabular}[c]{@{}l@{}}Y (brain region) = \\Age + Sex\\~\end{tabular} & \begin{tabular}[c]{@{}l@{}}HCP \& UKB, \\race is known \\but not included.\end{tabular} & \begin{tabular}[c]{@{}l@{}}HCP \& UKB (table \ref{tab:table2}), \\race is known~and used \\post-hoc to quantify bias.\end{tabular} \\ 
\hline
Race included & \begin{tabular}[c]{@{}l@{}}Y (brain region) = \\Age + Sex + Race\\~\end{tabular} & \begin{tabular}[c]{@{}l@{}}HCP \& UKB, \\race is known\\~and included.\end{tabular} & \begin{tabular}[c]{@{}l@{}}HCP \& UKB (table \ref{tab:table2}), \\race is known and used \\post-hoc to quantify bias.\end{tabular} \\
\hline
\end{tabular}
\end{table}

\subsection{Evaluating Fairness of Normative Models} 

\subsubsection{Qualitative evaluation: summarizing models for each racial group}

To evaluate the normative models in different racial groups, we calculated the average deviation score at every brain region for each group (Figure \ref{fig:figure2}A). We summarize patterns of extreme deviation ($|Z|>2$) for each group by counting how many subjects had an extreme deviation ($|Z|>2$) at a given brain region and dividing by the group size to show the percentage of individuals with extreme deviations at that brain area (Figure \ref{fig:figure2}B). 

\subsubsection{Quantitative evaluation: testing for group differences}

To test for statistically significant group difference in the normative models, we performed group difference testing on the deviation scores and residual errors, thresholding the results at a Benjamini-Hochberg \cite{hochberg_more_1990} false discovery rate (FDR) of $p < 0.05$ to correct for multiple comparisons across all brain regions (Figure \ref{fig:figure3}, Table \ref{tab:table4}).

\subsection{Predicting Race}

To further test if race is encoded in the normative models, we used deviation scores as features (separately for each normative model described in Table \ref{tab:table3}) to predict self-reported race using logistic regression one vs. rest framework (i.e., Black vs. White + Asian, Asian vs. White + Black, and White vs. Asian + Black). For this, we used penalized logistic regression implemented in scikit-learn \cite{pedregosa_scikit-learn_2011} with default settings (L2 penalty, L-BFGS solver). 

Data (Tables \ref{tab:table1} and \ref{tab:table2}) were divided into training and testing, stratified by race (so the test sets had approximately the same number of Asian, Black, White individuals), using an 80/20 split and generalization was assessed with 5-fold cross validation. Area under the curve, precision, recall, and F-score were determined for each class (Table \ref{tab:table5}), as well as visualizing the receiver operating characteristic curves and confusion matrices (Figure \ref{fig:figure4}). All metrics were averaged across folds and the mean and standard deviation are reported (Table \ref{tab:table5}). 

\subsection{Fairness Metrics}
We use the concept of demographic parity as our main fairness metric. Parity is often used in decision making binary classification settings (e.g., did you get the loan?) In these yes or no settings, there is a “positive” outcome (got the loan) and parity states that all groups should receive the positive outcome at equal rates. In this work, we have (i) a regression problem (the normative model) and (ii) a multi-class classification problem that directly predicts the sensitive attribute, thus there is no decision or positive outcome. Therefore, we use slightly different notions of parity in our assessments. We define parity as the normative model providing equal performance for all groups. Note that in the settings where this is not met, the resulting bias (e.g. a poorer fit in some groups than in others) will simply be propagated to the downstream analysis step. In the classification step, we define parity as equal classification performance across racial groups.  

\section{Results}

\subsection{Racial bias is present in normative models}

\subsubsection{Qualitative evaluation of racial bias}

\begin{figure}[!h]
  \centering 
  \includegraphics[width=\textwidth]{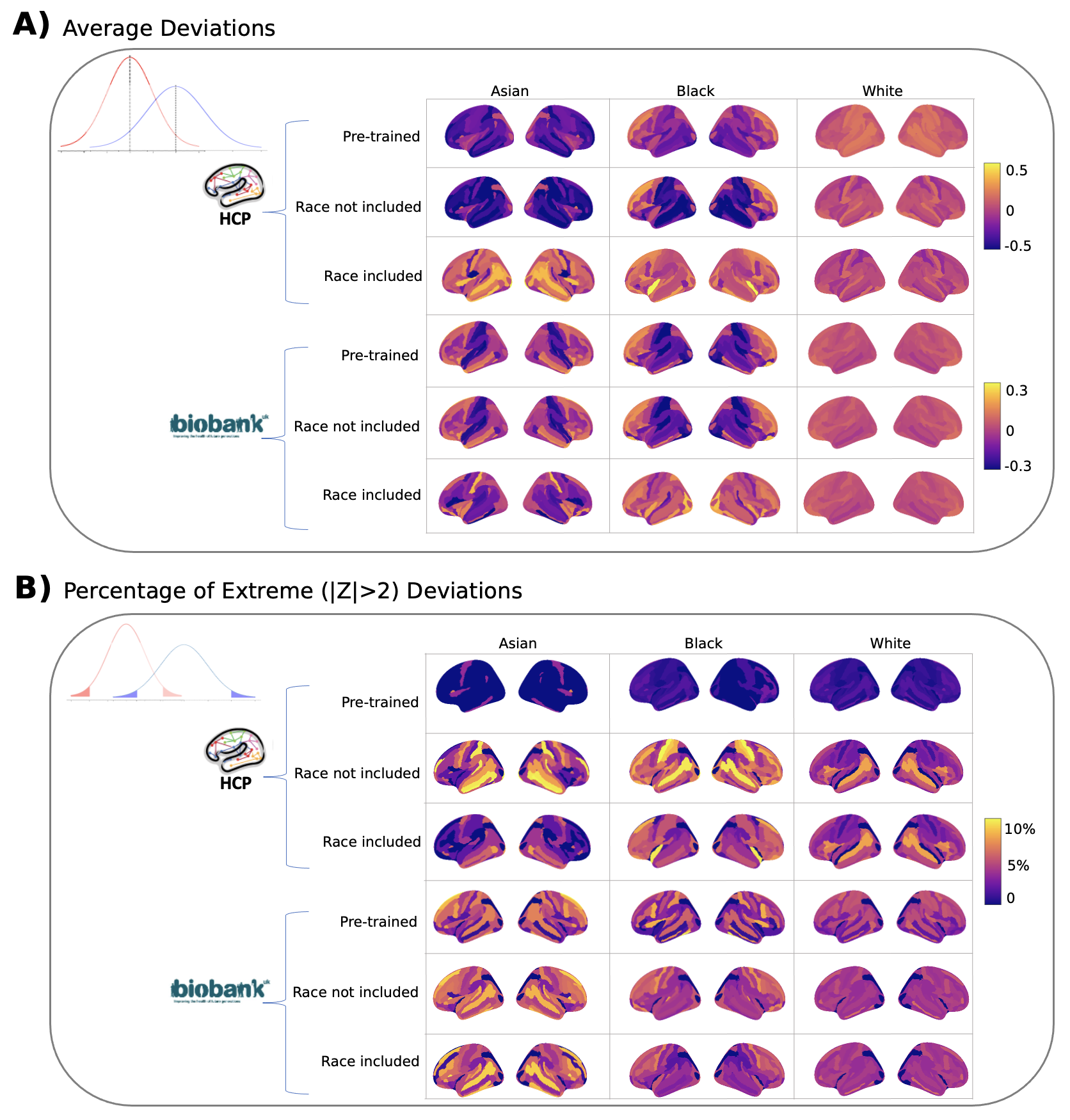} 
  \caption{Summary of normative model deviation scores across all three reference classes (pre-trained, race not included, and race included) in HCP and UKB datasets. A) Average (mean) deviations for all brain regions within all racial groups (columns). B) Percentage of extreme deviations (positive and negative) for all brain regions within all racial groups (columns).}
  \label{fig:figure2} 
\end{figure}

The average deviation summaries (Figure \ref{fig:figure2}A) measure the average deviation across each racial group. These show that white individuals (i.e. the majority class) are centered around zero across all reference class normative model (and across the whole brain) in both HCP and UKB datasets. The pre-trained average deviations show that for Asian individuals and Black individuals’ cortical thickness is overestimated, meaning there are negative deviation scores across most of the brain, with the exception of the prefrontal cortex in HCP Black individuals, UKB Black individuals, and UKB Asian individuals, and in the UKB Asian individuals’ temporal and visual areas. The ‘race not included’ average deviations follow a similar pattern as shown in the pre-trained average deviations, although there is somewhat stronger overestimation (i.e. more negative deviations) of Asian individuals and Black individuals in the race not included models. When race is included, the overestimation of HCP Asian individuals, HCP Black individuals, and UKB Black individuals disappears. This is replaced by an underestimation of cortical thickness in HCP Asian individuals, HCP Black individuals, and UKB Black individuals in temporal and visual areas, but mostly the models are centered around zero and these group differences, for the most part, disappear. In the UKB Asian individuals, an interesting phenomenon occurs where the patterns of average deviations in the pre-train and race not included seem to flip signs in the race included models. We interpret this as being because either (i) there are insufficient numbers of Asian individuals in the training set to get a good estimate, or (ii) encoding race using one-hot dummy variables does not provide sufficient flexibility to model race differences across demographic groups. In the pre-trained and race not included average deviations, UKB Asian individuals are overestimated in motor cortex, but not the rest of the brain. However, in the race included models, the motor cortex is underestimated, and the rest of the brain is underestimated. 

The extreme deviation summaries (Figure \ref{fig:figure2}B) measure the number of individuals in the tail of the normative distribution at each region and are therefore more sensitive to differences in the shape of the distribution rather than the mean. These show several interesting features. In HCP, the pre-trained normative models seem to be relatively fair in terms of identifying extreme deviations. There is a constant pattern of extreme deviations across all groups in the pre-trained models. One could interpret this as a particular type of parity, as noted above. Also in the HCP data, the number of extreme deviations in the White group does not change much between the race included and race not included models. In HCP Asian and Black groups, there are more extreme deviations in the race not included models than in the race included models. In UKB, across all three normative models, Asian individuals seems to have more extreme deviations than the Black and White individuals. In UKB Asian individuals, there is not much difference in the extreme deviations across normative models, but there are slightly more extreme deviations in the race included normative models. In UKB Black individuals, the most extreme deviations are seen in the pre-trained normative models and there is not much difference between race included and race not included extreme deviations. In UKB White individuals, there is also not much difference across normative models in the extreme deviations.

Taken together, these results indicate that the distribution of different racial groups have a different mean, and that the shape of the distribution is different.  

\subsubsection{Quantitative evaluation of racial bias}

\begin{figure}[!h]
  \centering 
  \includegraphics[width=5.3in]{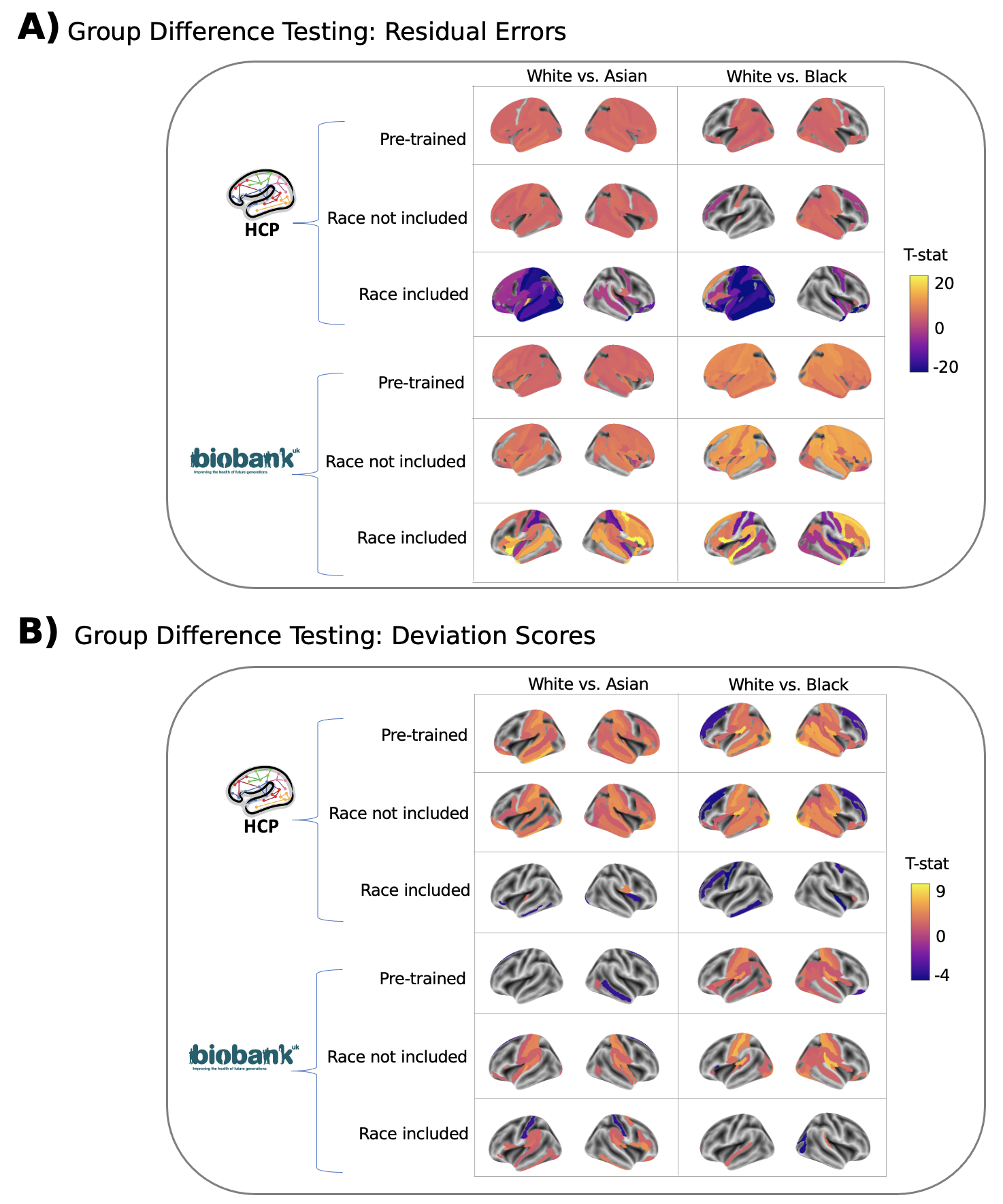} 
  \caption{Group differences in A) residual errors and B) deviation scores across all three reference classes (pre-trained, race not modeled, and race modeled) in HCP and UKB. The t-statistic is plotted where White individuals are group one, and Asian or Black individuals are group two. Light colors (positive t-stat) represent larger residual errors or deviations in White individuals and dark colors (negative t-stat) represent larger residual errors or deviations in Asian or Black individuals. Brain regions with statistically significant group differences after multiple comparison correction (FDRcorr $p<0.05$) are shown. The number of brain regions showing group differences for each model is shown in Table \ref{tab:table4}.}
  \label{fig:figure3} 
\end{figure} 

There were significant group differences in residual errors (equation \ref{eq:error}) (Figure \ref{fig:figure3}A) and deviation scores (equation \ref{eq:zscore}) (Figure \ref{fig:figure3}B) in all normative models after multiple comparison correction was performed (Table \ref{tab:table4}). The lowest percentage of group differences was consistently observed in the race included normative models, with the exception of HCP residual errors and UKB deviations. In HCP residual errors, White vs. Black showed the least differences in the race not included normative models and in UKB deviations, White vs. Asian showed the least differences in the pre-trained normative models. An interesting result is that the residual errors are higher in the White group in almost every model (Figure \ref{fig:figure3}A), except for HCP race included where there is a strong lateralization effect where the left hemisphere residual errors are greater in the Asian and Black groups than in the White group. UKB race included also shows a few regions where residual errors in the Asian group and in the Black group are greater than the White group residual errors. The residual errors being higher in the White group may be due to the much larger sample size (power) in this group. The direction of group differences, $White > Asian$  and $White > Black$, in the deviation scores (Figure \ref{fig:figure3}B) makes sense, as we show above in the average deviations (Figure \ref{fig:figure2}A) that the models are often underestimated for Asian and Black groups.

\begin{table}
\centering
\caption{Group differences in deviations and residual errors across all three reference classes (pre-trained, race not included, and race included) in HCP and UKB. We show the percentage of models with statistically significant differences after multiple comparison correction ($FDR_{corr} p<0.05$). In the Group column, W vs. A = White vs. Asian and W vs. B = White vs. Black.}
\label{tab:table4}
\begin{tabular}{c|c|c|c|c|c}
Dataset & Group & Metric & Pre-train & \begin{tabular}[c]{@{}c@{}}Race \\not included\end{tabular} & \begin{tabular}[c]{@{}c@{}}Race \\included\end{tabular} \\ 
\hline
HCP & W vs. A & deviations & 40\% & 49\% & 9\% \\
HCP & W vs. B & deviations & 55\% & 51\% & 5\% \\ 
\hline
UKB & W vs. A & deviations & 17\% & 25\% & 19\% \\
UKB & W vs. B & deviations & 45\% & 37\% & 7\% \\ 
\hline
HCP & W vs. A & error & 74\% & 64\% & 55\% \\
HCP & W vs. B & error & 56\% & 28\% & 54\% \\ 
\hline
UKB & W vs. A & error & 71\% & 56\% & 53\% \\
UKB & W vs. B & error & 87\% & 73\% & 51\% \\
\hline
\end{tabular}
\end{table}

\subsection{Race can be identified with high accuracy from normative models}

\begin{figure}[!h]
  \centering 
  \includegraphics[width=\textwidth]{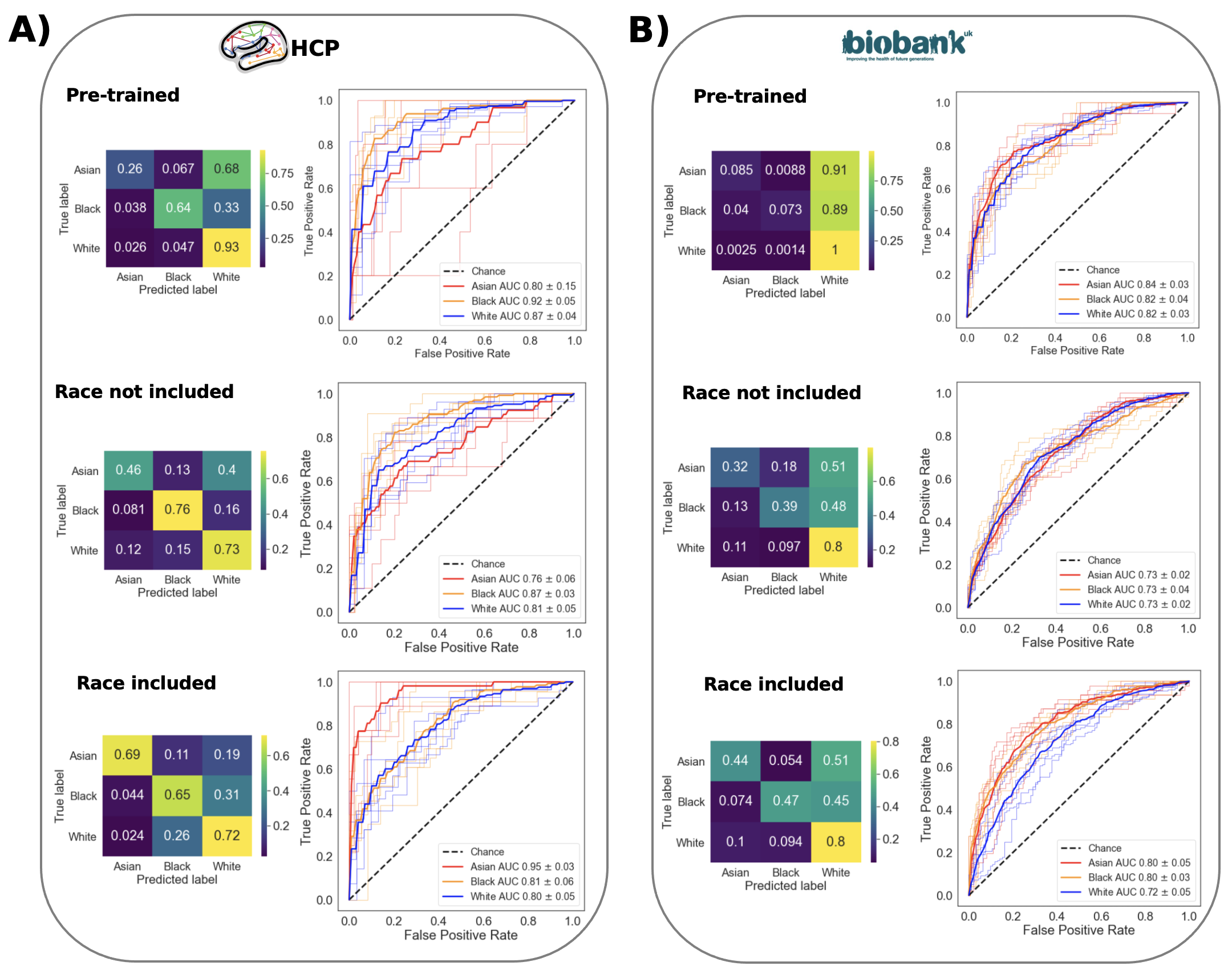} 
  \caption{Prediction of self-reported race in A) HCP and B) UKB datasets using deviation scores from three different reference class normative models (pre-trained, race not included, and race included) as features. Performance is evaluated with confusion matrices, receiver operator characteristic (ROC) curves, and Area under the ROC curve (AUC). For the confusion matrix interpretation, the diagonal elements show where predicted label == true label, and the off-diagonal elements show mislabeled (predicted label != true label). The confusion matrices were normalized by the true labels to show ratios rather than counts. For interpreting the receiver operator characteristic (ROC) curves, we plot the performance across 5-fold cross validation (lighter colors, thin lines) and the also the mean across all folds (darker colors, thicker lines).}
  \label{fig:figure4} 
\end{figure} 

In HCP, the pre-trained and race included models show similar average AUC and the race not included model had the lowest average AUC. White and Black individuals had the highest AUC in the pre-trained models and lowest AUC in the race included models. Asian individuals had the highest AUC in the race included and lowest in the race not included models. Pre-trained models have the lowest average precision, recall, and F-score but there is larger variance across racial groups. White individuals had the highest precision, recall, and F-score in the pre-trained models and lowest in the race included models. Black individuals had the highest precision, recall, and F-score in the race not included models, the lowest precision and F-score in the race included models, and the lowest recall in the pre-trained models. Asian individuals had the highest precision, recall, and F-score in the race included models and lowest in the pre-trained models.

In UKB, the pre-trained models showed the highest average and per group AUC. The race not included models had the lowest average, Asian group, and Black group AUC. The lowest AUC for the White group was in the race included models. As observed in HCP, the pre-trained models had the lowest average precision, recall, and F-score but increased variance across racial groups. White individuals had the highest precision, recall, and F-score in the pre-trained models and the lowest in the race not included models. Black and Asian individuals had the highest precision, recall, and F-score in the race included models and the lowest in the pre-trained models.

Taken together, these results show that even after accounting for differences across racial groups in the normative model estimation, sufficient information remains in the deviation scores to accurately predict self-reported race.

\begin{table}
\centering
\caption{Race prediction using normative model (pre-train, race not included, and race included) deviation scores as features. One vs Rest classification was used within 5-fold cross validation. Values shown are the average across all folds (\textpm s.d.). The mean across all race categories for each data set and model is shown in the mean (M) columns (mean of the column to the left). In the Group column A = Asian, B = Black and W = White.}
\label{tab:table5}
\setlength{\tabcolsep}{2.5pt}
\begin{tabular}{c|cc|cc|cc|cc}
\begin{tabular}[c]{@{}c@{}}\\\\\\\end{tabular} & Data & Group & Pre-train & M & \begin{tabular}[c]{@{}c@{}}Race \\not \\included\end{tabular} & M & \begin{tabular}[c]{@{}c@{}}Race \\included\end{tabular} & M \\ 
\hline
\multirow{6}{*}{AUC} & HCP & A & 0.80\textpm0.15 & \multirow{3}{*}{~0.86\textpm0.05} & 0.76\textpm0.06 & \multirow{3}{*}{~0.81\textpm0.04} & 0.95\textpm0.03 & \multirow{3}{*}{~0.85\textpm0.07} \\
 & HCP & B & 0.92\textpm0.05 &  & 0.87\textpm0.03 &  & 0.81\textpm0.06 &  \\
 & HCP & W & 0.87\textpm0.04 &  & 0.81\textpm0.05 &  & 0.80\textpm0.05 &  \\ 
\cline{2-9}
 & UKB & A & 0.84\textpm0.03 & \multirow{3}{*}{\begin{tabular}[c]{@{}c@{}}~0.83\textpm0.01\\~\end{tabular}} & 0.73\textpm0.02 & \multirow{3}{*}{~0.73\textpm0.0} & 0.80\textpm0.05 & \multirow{3}{*}{~0.77\textpm0.04} \\
 & UKB & B & 0.82\textpm0.04 &  & 0.73\textpm0.04 &  & 0.80\textpm0.03 &  \\
 & UKB & W & 0.82\textpm0.03 &  & 0.73\textpm0.02 &  & 0.72\textpm0.05 &  \\ 
\hline
\multirow{6}{*}{\begin{tabular}[c]{@{}c@{}}\\Precision \\\end{tabular}} & HCP & A & 0.31\textpm0.20 & \multirow{3}{*}{~0.64\textpm0.24} & 0.43\textpm0.10 & \multirow{3}{*}{~0.64\textpm0.15} & 0.79\textpm0.16 & \multirow{3}{*}{~0.71\textpm0.06} \\
 & HCP & B & 0.70\textpm0.04 &  & 0.76\textpm0.07 &  & 0.65\textpm0.09 &  \\
 & HCP & W & 0.90\textpm0.02 &  & 0.74\textpm0.09 &  & 0.70\textpm0.08 &  \\ 
\cline{2-9}
 & UKB & A & 0.21\textpm0.08 & \multirow{3}{*}{~0.32\textpm0.33} & 0.40\textpm0.03 & \multirow{3}{*}{~0.52\textpm0.14} & 0.52\textpm0.11 & \multirow{3}{*}{~0.61\textpm0.09} \\
 & UKB & B & 0.35\textpm0.34 &  & 0.45\textpm0.05 &  & 0.57\textpm0.07 &  \\
 & UKB & W & 0.98\textpm0.0 &  & 0.72\textpm0.02 &  & 0.73\textpm0.03 &  \\ 
\hline
\multirow{6}{*}{Recall~} & HCP & A & 0.26\textpm0.23 & \multirow{3}{*}{~0.61\textpm0.27} & 0.46\textpm0.15 & \multirow{3}{*}{~0.65\textpm0.13} & 0.69\textpm0.14 & \multirow{3}{*}{~0.69\textpm0.03} \\
 & HCP & B & 0.63\textpm0.08 &  & 0.76\textpm0.07 &  & 0.65\textpm0.12 &  \\
 & HCP & W & 0.93\textpm0.02 &  & 0.73\textpm0.08 &  & 0.72\textpm0.09 &  \\ 
\cline{2-9}
 & UKB & A & 0.09\textpm0.04 & \multirow{3}{*}{~0.38\textpm0.43} & 0.32\textpm0.04 & \multirow{3}{*}{~0.50\textpm0.21} & 0.44\textpm0.10 & \multirow{3}{*}{~0.57\textpm0.17} \\
 & UKB & B & 0.07\textpm0.06 &  & 0.39\textpm0.05 &  & 0.47\textpm0.05 &  \\
 & UKB & W & 0.99\textpm0.0 &  & 0.80\textpm0.03 &  & 0.81\textpm0.05 &  \\ 
\hline
\multirow{6}{*}{F-score} & HCP & A & 0.27\textpm0.20 & \multirow{3}{*}{~0.61\textpm0.26} & 0.44\textpm0.12 & \multirow{3}{*}{\begin{tabular}[c]{@{}c@{}}~0.64\textpm0.14\\~\end{tabular}} & 0.74\textpm0.14 & \multirow{3}{*}{~0.70\textpm0.04} \\
 & HCP & B & 0.66\textpm0.05 &  & 0.76\textpm0.06 &  & 0.64\textpm0.09 &  \\
 & HCP & W & 0.91\textpm0.01 &  & 0.73\textpm0.08 &  & 0.71\textpm0.07 &  \\ 
\cline{2-9}
 & UKB & A & 0.12\textpm0.05 & \multirow{3}{*}{~0.41\textpm0.41} & 0.35\textpm0.03 & \multirow{3}{*}{~0.51\textpm0.18} & 0.47\textpm0.10 & \multirow{3}{*}{~0.58\textpm0.13} \\
 & UKB & B & 0.12\textpm0.10 &  & 0.42\textpm0.05 &  & 0.52\textpm0.05 &  \\
 & UKB & W & 0.99\textpm0.0 &  & 0.76\textpm0.02 &  & 0.76\textpm0.03 & 
\end{tabular}
\end{table}

\section{Discussion} 

In this study, we used two independent datasets to document the racial bias present in reference class normative models of neuroimaging-derived brain structure. Across three different normative models, we uncovered differences in the way deviations were identified across racial groups. Racial bias was still present even when including race as a predictor in the normative model. This is an important finding because (i) it suggests that the modeling approach used here is not sufficiently flexible to account for racial differences, and (ii) deviations from the reference class “norm” (average/mean) are often interpreted with clinical meaning (i.e., representing biological dysfunction) when in reality we show that deviations could be due to an individual’s demographic mismatch with the reference class.  However, including race as a covariate in normative models did increase the parity of classifier predictions. We also show that self-reported race can be accurately predicted from normative model features well above chance. This discovery is surprising because this is considered to be a challenging task for human experts (i.e., a neuroradiologist could not easily identify a patient’s race from looking at an MRI of their brain).

Identifying self-reported race from neuroimaging-derived features is not necessarily a meaningful goal on its own. Our results in the normative model setting align with work using deep learning models \cite{gichoya_ai_2022,seyyed-kalantari_underdiagnosis_2021} to predict race directly from medical images (i.e., not reference class normalized images) to show that racial information cannot easily be isolated or removed from imaging data. The purpose of this work is not to point out new cases of racial disparities in model performance. Rather, it is important to view this within the broader context of employing predictive models in real-world clinical scenarios (where there are known racial disparities \cite{adam_write_2022,seyyed-kalantari_underdiagnosis_2021,xiao_name_2024,chang_disparate_2022,pierson_algorithmic_2021,li_cross-ethnicityrace_2022} and recognize that these models are not colorblind \cite{watson-daniels_algorithmic_2024}. When predicting race based on medical images or other healthcare data, it's essential to ask: What exactly are we predicting? Are we predicting race as in someone's skin color? Or (more likely) are we predicting an identity intertwined with unmeasured risk factors or exposures, including the sociohistorical pressures (individual and structural racism) that shape group differences. We cannot ignore the fact that racism exists and is coded in technologies. This contemplation naturally segues into a discussion about race and genetic ancestry. Race is a social construct, categorizing individuals based on perceived physical traits, while genetic ancestry is contextualized within reference populations, indicating the genetic heritage of a population \cite{borrell_race_2021,yudell_taking_2016,maglo_population_2016}. Differences in clinical measures and outcomes among racial groups are frequently depicted as inherent biological distinctions \cite{vyas_hidden_2020}. However, these differences are typically studied using self-reported race (social construct) not genetic ancestry \cite{vyas_challenging_2019,kowalsky_case_2020}. \textbf{Given this knowledge, it is critical to acknowledge that the racial group differences we observe in this work cannot be interpreted biologically.}

Our results show that the linear approach we use to modeling race is not sufficient to accommodate racial differences. While it is likely that these can be more adequately modeled using non-linear techniques such as neural networks, it is crucial to recognize that – at least in brain imaging – we lack sufficiently representative cohorts to be able to estimate a reference class that faithfully reflects variation across different demographic groups. While pooling of neuroimaging datasets is becoming more prevalent in the field \cite{rutherford_charting_2022,bethlehem_brain_2022}, most legacy datasets do not contain racial information making it impossible to even assess racial bias in these models. While more recent cohort studies such as UKB and HCP do acquire this information, these samples are still heavily biased toward WEIRD (Western, Educated, Industrialized, Rich, Democratic individuals \cite{henrich_weirdest_2010} (as shown in Table \ref{tab:table1} and \ref{tab:table2}). Therefore, acquiring more representative cohorts should be seen as an urgent research priority.

\paragraph{Limitations} There are several limitations of this work concerning the data that was used. Compared to other machine learning in healthcare settings that use different types of data such as electronic medical records (e.g., the MIMIC dataset) that contain ~425,000 patient visits \cite{johnson_mimic-iv-ed_2023}, the sample size used in this study is rather modest (Table \ref{tab:table1} and \ref{tab:table2}) but it is in line with sample sizes used in neuroimaging studies \cite{greene_brainphenotype_2022,rosenberg_how_2022}. Regarding the representation of the neuroimaging data, we used features that were extracted from a group average template space that is defined based on a non-diverse sample (129 White individuals, 15 Asian individuals, and 1 person of mixed race) \cite{fonov_unbiased_2009}. Mapping individual’s neuroimaging data into this space may create or conceal structural brain differences across individuals. Regarding the demographic variables, we only considered coarse race labels (Asian, Black, White) due to data availability. There is a need to question how race is defined in the algorithmic fairness community and how this impacts the utility of algorithmic fairness in real-world equity goals \cite{abdu_empirical_2023}. Recent work has also shown in the performance metrics of several clinical prediction tasks, there is greater variability within (coarse) racial group labels than between (coarse) racial groups \cite{movva_coarse_2023}. Thus, considering more granular racial categories would likely help the interpretation of our group difference findings. In addition to including more granular racial groups, intersectionality of sensitive groups is also an important (e.g., Black woman) consideration \cite{tolbert_correcting_2023} as well as including socioeconomic indicators such as education and income \cite{yang_evaluating_2023}.

\paragraph{Conclusion} It is not sufficient to model all racial groups in the same reference class, given the current frequencies available in the training data where there is over-representation of White individuals. We observed racial disparities in the normative models, in the univariate (group difference testing) and multivariate (race prediction) sense, that likely arise due to lack of data diversity. We reiterate that these findings cannot be interpreted biologically. Rather, race is a proxy for the multifaceted interactions among genetic ancestry, race, racism, socioeconomic status, and other environmental factors. These factors have not been well documented in existing datasets. In other words, we need to collect not just bigger data, but more diverse and representative data \cite{kopal_end_2023} which has been recognized with research initiatives such as the \hyperlink{https://commonfund.nih.gov/bridge2ai}{NIHBridge2AI}. This should include granular measures of race and ethnicity, as well as additional information on experiences of racism, socioeconomic status, and environmental factors in order to begin disentangling the mechanisms of heterogeneity \cite{carter_its_2022, cardenas-iniguez_we_2023, yang_evaluating_2023}.

Transparency should be a priority, given it is unlikely to achieve perfectly fair, or unbiased predictive models. This means communicating known biases (i.e., well documented sample demographics) of the training, validation, and testing sets. There are existing tools to help with this communication, including datasheets for datasets \cite{gebru_datasheets_2021} and model cards for model reporting \cite{mitchell_model_2019}. Subgroup performance audits are a key first step in revealing underlying issues that should be addressed before model integration. \textbf{Due to unmeasured dimensions of bias, acknowledging that predicted phenotypes reflect complex combinations of variables should also be part of routine model interpretation.}

A recent editorial on biased machine learning in healthcare shifts the fairness framework from viewing biased health data solely as detrimental to considering it as informative artifacts. The authors propose that AI’s powerful pattern recognition abilities can be leveraged to detect exclusion from preventive healthcare and can serve as a hypothesis-generating tool that motivates new research on health inequities in healthcare \cite{drazen_considering_2023}. We agree with this change of perspective on biased data and models. While these results reveal uncomfortable truths about our data (and societies in which data was collected), they also align with public health goals regarding health equity \cite{lin_development_2024} which recognizes that acknowledging and understanding population inequities is an essential first step to making progress.

\acks{The authors thank Konstantin Genin for contributing to fruitful conversations about the reference class problem in the context of normative modeling. This research was supported by grants from the European Research Council 10100118 (AFM), the Dutch Organisation for Scientific Research VIDI grant 016.156.415 (AFM), the Harvard Catalyst Program for Diversity and Inclusion (NGH), Brain Behavior Research Foundation Young Investigator Award (NGH), National Institutes for Mental Health K01MH129828 (NGH), and German Research Foundation Emmy Noether 513851350 (TW).}

\bibliography{DemoNM}

\newpage
\appendix
\section*{Appendix A.}

\begin{figure}[!h]
  \centering 
  \includegraphics[width=\textwidth]{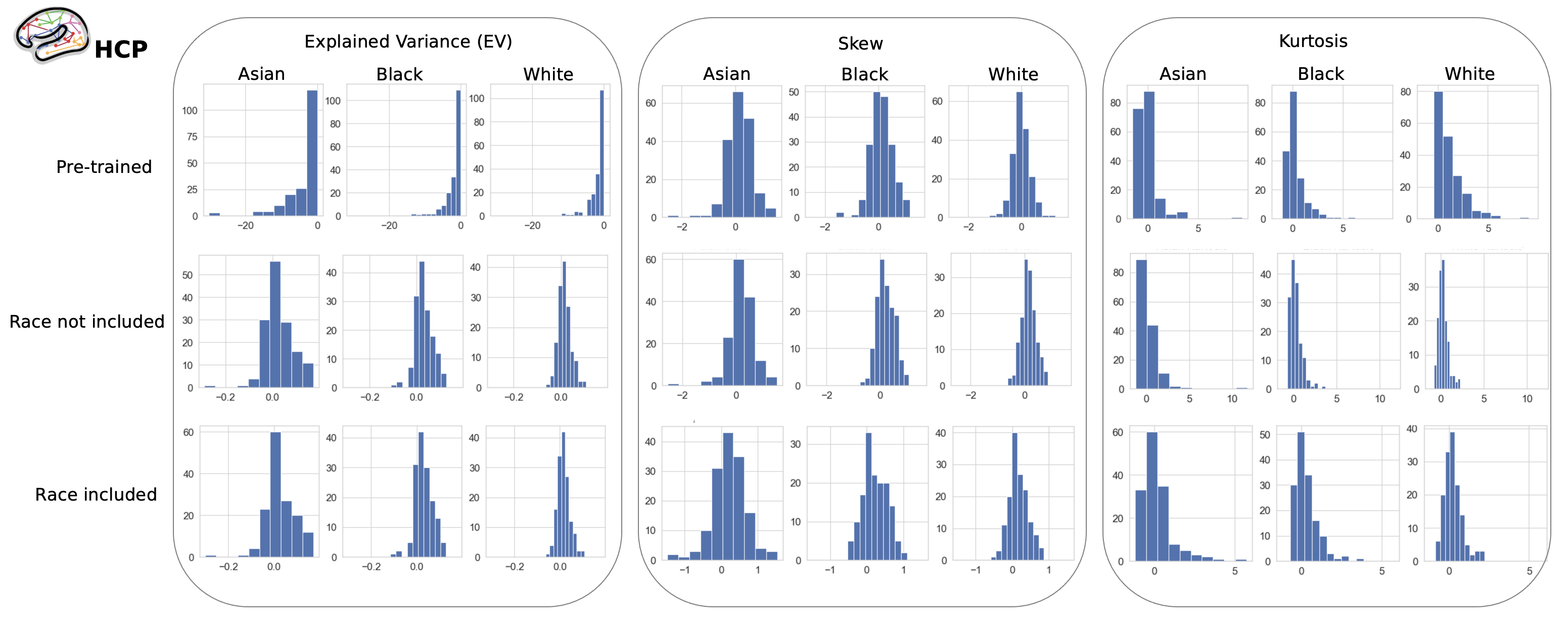} 
  \caption{Evaluation metrics for normative models in HCP dataset.}
  \label{fig:supfig1} 
\end{figure} 

\begin{figure}[!h]
  \centering 
  \includegraphics[width=\textwidth]{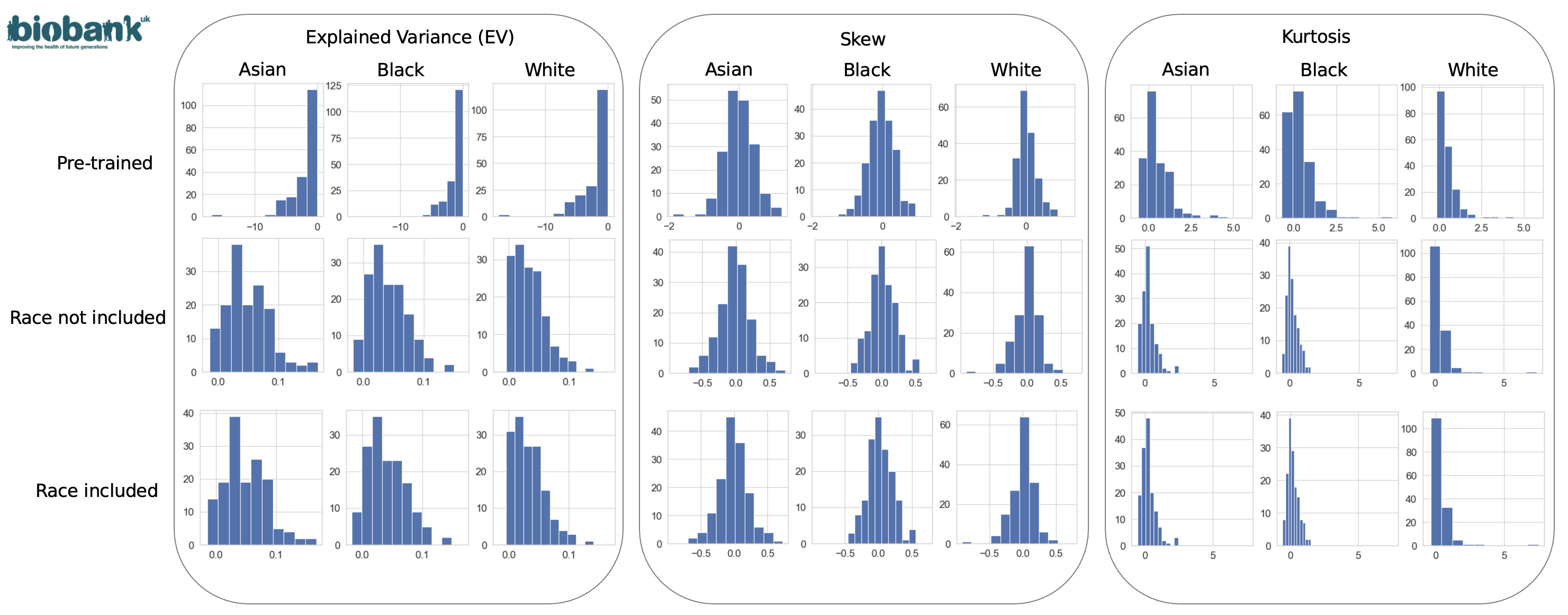} 
  \caption{Evaluation metrics for normative models in UKB dataset.}
  \label{fig:supfig2} 
\end{figure} 

\end{document}